\documentclass[letterpaper, 10 pt, conference]{ieeeconf}  

\IEEEoverridecommandlockouts                              

\usepackage{amsmath} 
\usepackage{amssymb}  
\usepackage{graphicx}
\usepackage{booktabs}
\usepackage{xcolor}
\usepackage{makecell}
\definecolor{myLinkColor}{HTML}{1A73E8}
\usepackage[colorlinks=true, urlcolor=myLinkColor]{hyperref}
\usepackage[skip=3pt, font={small}]{caption}
\usepackage{bm}
\usepackage{cases}
\usepackage{balance}

\title{\LARGE \bf
HITTER: A HumanoId Table TEnnis Robot \\ via Hierarchical Planning and Learning
}

\author{Zhi Su, Bike Zhang, Nima Rahmanian, Yuman Gao, Qiayuan Liao, Caitlin Regan,\\
Koushil Sreenath, S. Shankar Sastry 
\thanks{The authors are with the University of California, Berkeley.}
}

\makeatletter
\let\@oldmaketitle\@maketitle
\renewcommand{\@maketitle}{\@oldmaketitle
  \vspace{0.3cm}
  \centering
  \setcounter{figure}{0}%
  \begin{minipage}{\linewidth}
    \includegraphics[width=\textwidth]{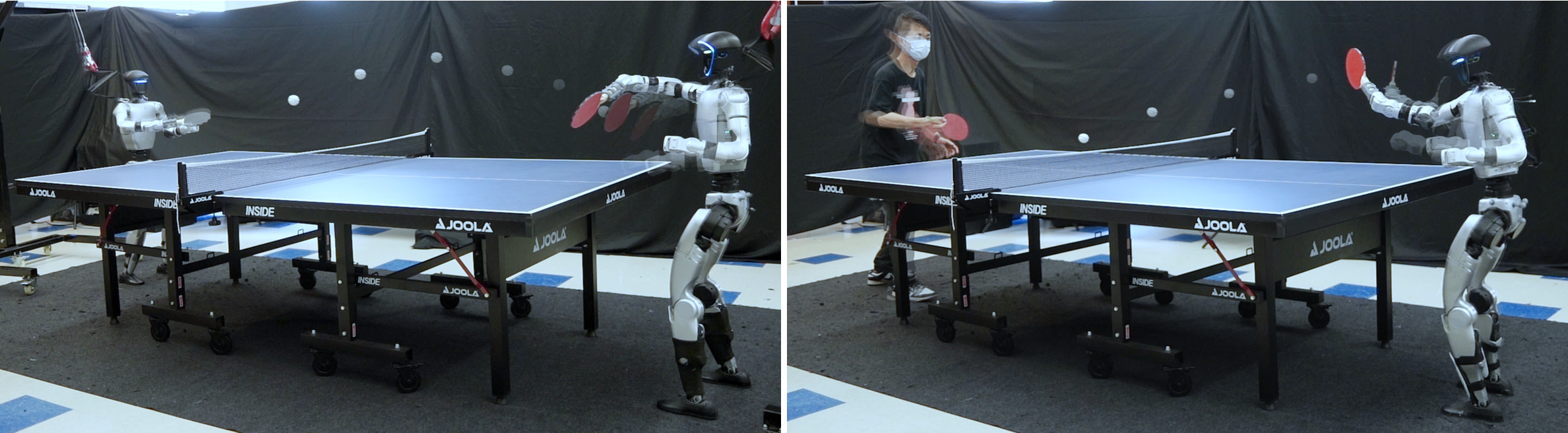}
    \vspace{0pt}
    {\captionsetup{hypcap=false}%
    \captionof{figure}{\label{fig:teaser}\textbf{Humanoid table tennis rallies.}
    Our system enables both humanoid-humanoid (left) and humanoid-human (right)
    matches, achieving rallies of up to 106 consecutive shots against a human
    opponent. Project website: \href{https://humanoid-table-tennis.github.io}{humanoid-table-tennis.github.io}.}}
  \end{minipage}
  \vspace{0pt}
}
\makeatother

\begin{document}

\maketitle

\thispagestyle{empty}
\pagestyle{empty}

\begin{abstract}
Humanoid robots have recently achieved impressive progress in locomotion and whole-body control, yet they remain constrained in tasks that demand rapid interaction with dynamic environments through manipulation. Table tennis exemplifies such a challenge: with ball speeds exceeding 5\,m/s, players must perceive, predict, and act within sub-second reaction times, requiring both agility and precision. To address this, we present a hierarchical framework for humanoid table tennis that integrates a model-based planner for ball trajectory prediction and racket target planning with a reinforcement learning-based whole-body controller. The planner determines striking position, velocity and timing, while the controller generates coordinated arm and leg motions that mimic human strikes and maintain stability and agility across consecutive rallies. Moreover, to encourage natural movements, human motion references are incorporated during training. We validate our system on a general-purpose humanoid robot, achieving up to 106 consecutive shots with a human opponent and sustained exchanges against another humanoid. These results demonstrate real-world humanoid table tennis with sub-second reactive control, marking a step toward agile and interactive humanoid behaviors.

\end{abstract}

\section{Introduction}
Humanoid robots have long been envisioned as general-purpose embodied intelligent agents capable of performing versatile tasks, that are anthromorphically compatible with our own actions. Recent progress has led to impressive demonstrations in locomotion~\cite{long2024learning} and motion imitation~\cite{liao2025beyond, xie2025kungfubot}. Nevertheless, most existing work remains focused on control in the free space or with static ground interaction. For example, maintaining balance~\cite{zhang2025hub}, walking~\cite{long2024learning}, or mimicking human poses~\cite{liao2025beyond}, while relatively few address the challenge of interacting with fast-moving objects in dynamic environments.

\begin{figure*}
    \centering
    \includegraphics[width=\linewidth]{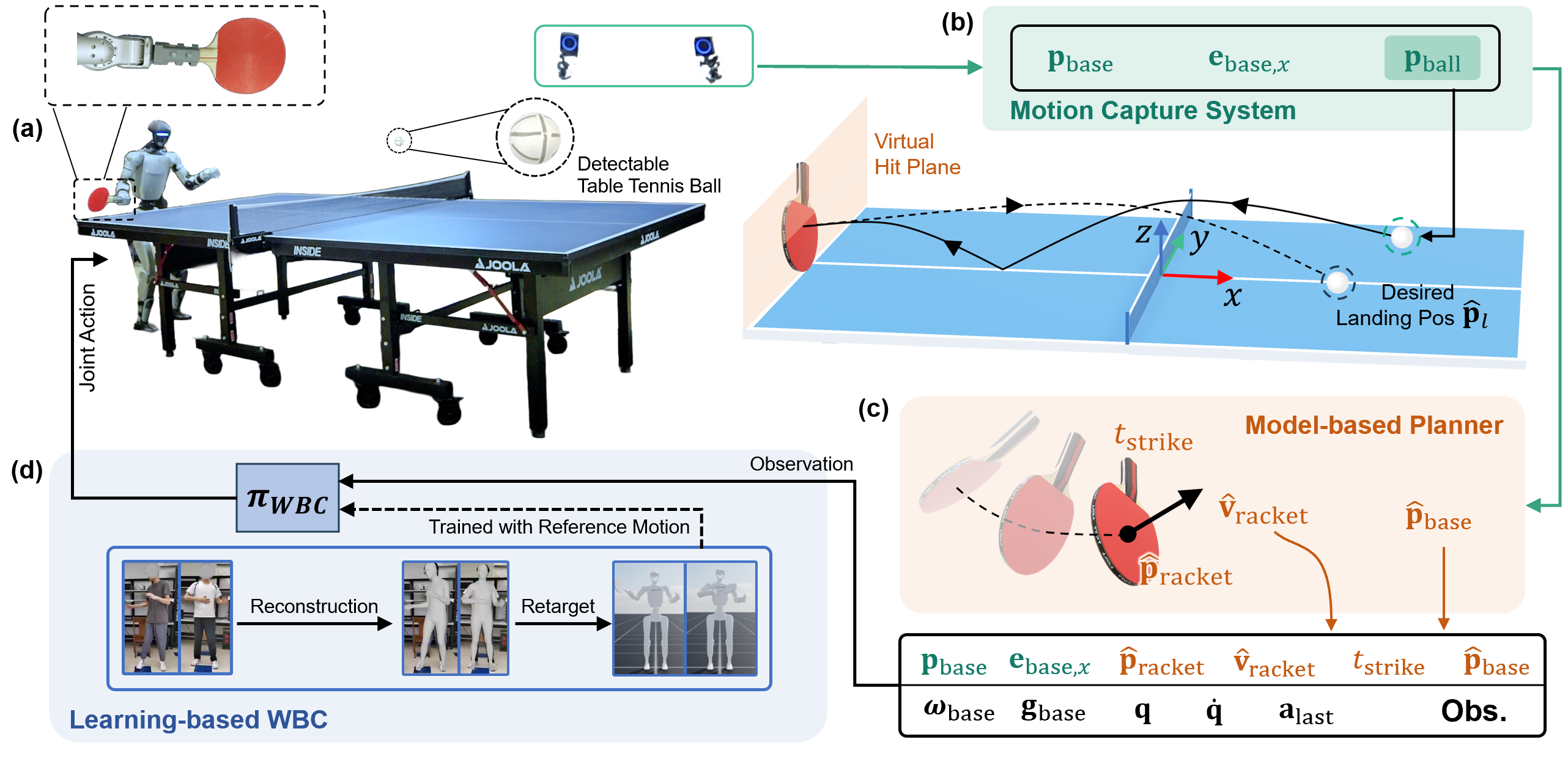}
    \caption{\textbf{System overview.} 
    (a) The racket is mounted on the robot’s right wrist using a 3D-printed connector, and the ball is covered with reflective tape for motion capture. 
    (b) The motion capture system tracks the ball position $\mathbf{p}_{\mathrm{ball}}$, the robot base position $\mathbf{p}_{\mathrm{base}}$, and the base forward vector $\mathbf{e}_{\mathrm{base},x}$. 
    (c) The model-based planner uses $\mathbf{p}_{\mathrm{ball}}$ and the desired landing point $\hat{\mathbf{p}}_l$ to predict the racket’s striking position $\hat{\mathbf{p}}_{\mathrm{racket}}$, velocity $\hat{\mathbf{v}}_{\mathrm{racket}}$, and strike time $t_{\mathrm{strike}}$. Given $\mathbf{p}_{\mathrm{base}}$ and $\hat{\mathbf{p}}_{\mathrm{racket}}$, the base target position $\hat{\mathbf{p}}_{\mathrm{base}}$ is also computed. 
    (d) The learning-based Whole Body Controller (WBC) policy $\pi_{WBC}$ is trained in simulation via reinforcement learning with human motion references and then deployed on the real robot. It takes as input the observations provided by other system components together with the robot’s proprioceptive information.}
    \label{fig:method}
\end{figure*}

Such interactions are fundamentally harder: they demand not only coordinated control across dozens of joints, but also tight perception-action loops operating at extreme time scales. Table tennis epitomizes this challenge. A ball traveling at over 5\,m/s allows for only sub-second reaction times, forcing the robot system to perceive, predict, plan, and strike within a few hundred milliseconds. Unlike locomotion or static manipulation, success requires agile whole-body movements, blending rapid arm swings with waist rotations, as well as quick stepping and balance recovery, to ensure accurate hit and readiness for the next rally. Compared to other racket sports, such as badminton~\cite{ma2025learning} or tennis~\cite{zaidi2023athletic}, table tennis poses an even greater challenge due to its shorter distances, faster exchanges, and smaller reaction windows. Humanoid table tennis therefore serves as a unique testbed for robotics because:
(a) it is highly dynamic, requiring fast ball trajectory prediction and strike planning;
(b) consecutive rallies necessitate agile motions and balance recovery after each stroke; and
(c) human-like striking motions are essential for both effectiveness and naturalness.

To address these challenges, we adopt a hierarchical framework that separates high-level planning from low-level control. At the high level, a model-based planner estimates the ball trajectory and predicts the striking position, velocity, and timing with high precision. At the low level, a reinforcement learning (RL)-based whole-body controller is trained to execute human-like striking motions while tracking the racket targets specified by the planner. This design directly addresses the identified challenges by: (a) having the planner provide fast trajectory prediction and strike planning, thereby offering a stable interface for the controller and improving sample efficiency during training; (b) training the RL controller on consecutive strikes so that it learns agile motions and reliable balance recovery; and (c) incorporating only two human reference motions into training, through which the controller acquires natural and human-like strikes.

Through extensive real-world experiments, including rallies against both humans and humanoids, we demonstrate that humanoid table tennis is feasible, marking a step toward agile and interactive humanoid behaviors. This is achieved using a general-purpose humanoid robot, without relying on specialized hardware. Notably, the system operates fully autonomously, without teleoperation. Our contributions are as follows:
\begin{itemize}
    \item We propose a hierarchical framework that integrates a model-based planner and an RL-based whole-body controller, enabling humanoid table tennis play.
    \item We develop human-like, whole-body striking skills, combining coordinated arm motions for multi-skill ball-hitting, and agile leg movements for rapid reaching.
    \item We validate our approach through extensive real-world experiments, including both humanoid-humanoid and humanoid-human rallies. Our humanoid achieves up to 106 consecutive shots against a human opponent.
\end{itemize}

\section{Related Work}

\subsection{Robotic Table Tennis}
Robotic table tennis has emerged as a rich testbed for high-speed perception, planning, and control, requiring rapid reactivity to return balls at competitive speeds.  In our own work, we were very much inspired by the  biologically-inspired approach to motion generation in \cite{mulling2010biomimetic, mulling2011biomimetic}. In this work the research team led by Jan Peters fits a template trajectory while minimizing the robot's deviations from a fixed ``comfort posture": Human demonstration via kinesthetic teach-in was used in \cite{mulling2013generalizestrike} to learn  a general mixture of motions policy from motion primitives. 
The interest in responding to adversaries  with  opponent modeling was presented in \cite{Wang_Boularias_Mülling_Peters_2011} with  anticipatory action selection \cite{6094892}, and  the design of the Intention-driven Dynamics Model \cite{Wang2013} and \cite{WANG2017399}. The work in \cite{MullingBMSP2014} applied inverse reinforcement learning to identify strategic elements of play from demonstrations between different players. Our own work has analyzed detailed play from hours of recorded video \cite{etaat2025latte} to determine intent. In  Section \ref{disc-sec}, we briefly discuss integrating this into our humanoid going forward. 

Recent progress has combined advances in hardware, control, and learning. For instance, a lightweight, high-torque robotic arm with model predictive control was developed to execute diverse hit styles with precision \cite{nguyen2025high}. Similarly, a model-based approach for hitting velocity control has demonstrated high success rates in returning balls accurately \cite{ji2021model}. On the learning side, seminal work on the use of fast learning-based techniques for playing table tennis have been presented by the Gemini Robotics team at DeepMind \cite{Sanketi2023, amor2025sas}. In continuation of this work, amateur human-level competitive performance was achieved through a hierarchical architecture with sim-to-real adaptation \cite{d2024achieving}. In contrast to large-scale training efforts, \cite{tebbe2021sample} showed that  reinforcement learning can learning to rally from scratch in under 200 trials. Moreover, beyond specialized robotic table tennis devices, general-purpose humanoid robots have been shown to play table tennis using impedance control \cite{xiong2012impedance}. However, in prior work, the humanoid was constrained to stand still without agile locomotion, which limited the effective hitting range. In our work, we focus on human-like and agile reaching motions for table tennis hitting, a capability not previously demonstrated on humanoid robots.

\subsection{Humanoid Whole-Body Control}
Whole-body humanoid control has seen rapid advances in recent years.
Early efforts mainly focused on training humanoid robots to mimic specific human motions using large-scale reinforcement learning \cite{cheng2024expressive, he2024omnih2o}. Other approaches decouple control into separate upper- and lower-body policies to simplify learning and coordination \cite{zhang2025falcon}. 
More recent research has shifted toward developing general whole-body motion trackers capable of imitating a variety of human movements \cite{ze2025twist, chen2025gmt}.
Building upon these advances \cite{liao2025beyond}, our work introduces a humanoid whole-body controller specifically designed for rapid and dynamic interactions with the environment, such as ball hitting.
\section{System Overview}
We adopt a hierarchical system framework that integrates a model-based planner with a learning-based whole-body controller (Fig.~\ref{fig:method}).
Nine OptiTrack cameras track the ball’s position, with the motion capture system operating at 360\,Hz and achieving millimeter-level accuracy.
The estimated ball position is provided to the model-based planner, which predicts the hitting position and time, and computes the desired racket velocity at impact for the Unitree G1 humanoid~\cite{unitree_g1}.
The robot is equipped with a reinforcement learning policy that processes the observations and outputs the desired joint positions for all 29 joints at 50\,Hz.
These joint position setpoints are converted into joint torques using a PD controller.

As shown in Fig.~\ref{fig:method}, we use a regulation-size table measuring 2.74\,m by 1.525\,m, with the playing surface 0.76\,m above the ground. The coordinate frame is defined at the table center on the top surface, with the $x$-axis aligned with the table’s long side and the $z$-axis pointing upward. In the sections below, we assume the humanoid is positioned on the $x<0$ side of the table.

\section{Model-based Planner}
The model-based planner receives the ball’s position at each timestep and predicts the desired racket’s striking position, velocity, and timing. These predictions are then passed to the whole-body controller to generate robot motions.

\subsection{State Estimation}
\label{subsec:state_estimation}
Before predicting the ball’s trajectory, we first estimate its velocity, which cannot be directly obtained from the motion capture system. To do this, we perform a least-squares fit of a second-order polynomial to the ball’s position over time in each coordinate direction, $\mathbf{p}(t)=[p_x(t), p_y(t), p_z(t)]$. The fitting is performed using the nearest 31 position measurements, providing a smooth estimate of both position and velocity. Upon detecting a bounce on the table, we clear the position measurement buffer to prevent the inclusion of pre-bounce data. The smoothed value of the ball's position and velocity at the current timestep are then attained by evaluating $\mathbf{p}(t)$ and its derivative $\dot{\mathbf{p}}(t)$.

\subsection{Ball Trajectory Prediction}
To predict the ball trajectory, we adopt the same hybrid dynamics model as in~\cite{nguyen2025high}:

\begin{subnumcases}{\Sigma:\ }
  \mathbf{a} = -k \|\mathbf{v}\|\, \mathbf{v} + \mathbf{g}, & $\mathbf{p} \notin \mathcal{S}$, \label{eq:flight_dynamics}\\[6pt]
  \mathbf{v}^{+} = \mathbf{C}\,\mathbf{v}^{-},                        & $\mathbf{p} \in \mathcal{S}$,  \label{eq:bouncing_dynamics}
\end{subnumcases}
where~\eqref{eq:flight_dynamics} describes the continuous-time flight dynamics of the ball, with $\mathbf{a}$ and $\mathbf{v}$ denoting its acceleration and velocity, respectively. 
Here, $k$ is the aerodynamic drag coefficient and $\mathbf{g}$ is the gravity vector. 
We assume the spin is sufficiently small so that spin-induced effects, such as the Magnus force, can be neglected. Equation~\eqref{eq:bouncing_dynamics} models the discrete-time impact dynamics upon bouncing on the table, where $\mathbf{v}^-$ and $\mathbf{v}^+$ represent the pre- and post-impact velocities. 
The restitution matrix is
\begin{equation}
    \mathbf{C} = \mathrm{diag}(C_h,\, C_h,\, -C_v),
\end{equation}
with $C_h$ and $C_v$ denoting the horizontal and vertical restitution coefficients, respectively. The set of impact states is
\begin{equation}
    \mathcal{S} = \bigl\{\, \mathbf{p} = [p_x,\,p_y,\,p_z] \;\big|\; p_z = 0 \,\bigr\}.
\end{equation}

\begin{figure}
    \centering
    \includegraphics[width=1\linewidth]{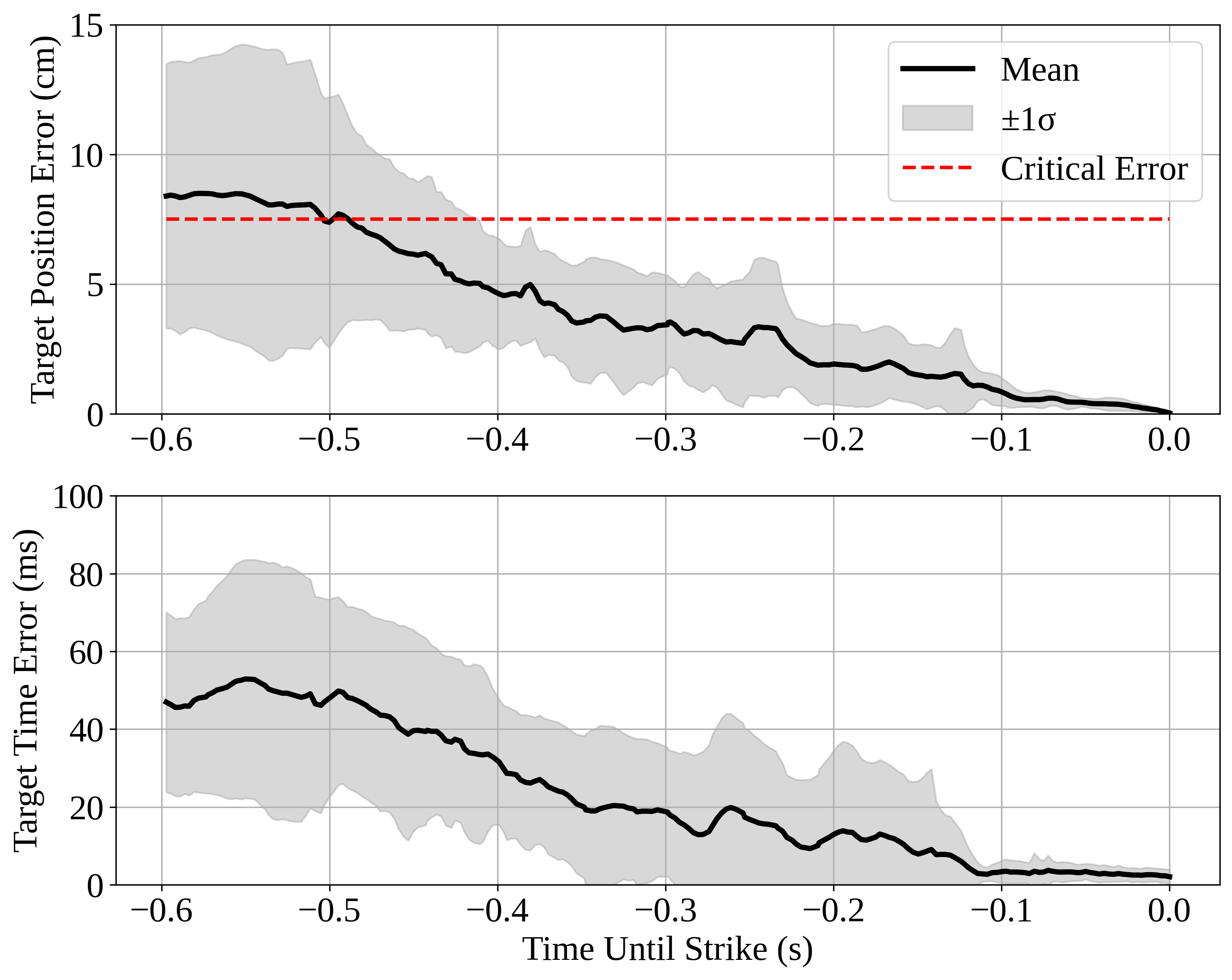}
    \caption{\textbf{Prediction errors of the model-based planner.} Striking position error (top) and striking time error (bottom) are evaluated over 20 ball trajectories. The shaded regions indicate the standard deviation, and the red dashed line marks the critical position error of 7.5\,cm, corresponding to the racket radius. At 0.5\,s before the strike, the position error falls below this threshold.}
    \label{fig:prediction_error}
\end{figure}

We estimate the parameters $k$, $C_h$, and $C_v$ from 15 recorded ball trajectories. 
In each trajectory, the ball is launched from one side of the table, bounces once on the opposite side, and then exits the table. 
The parameters are identified by fitting the recorded data to the following:
\begin{equation}
\begin{cases}
\|\mathbf{a}-\mathbf{g}\| = k\|v\|^2, \\
\{\|v_x^+\|, \|v_y^+\|\} = C_h \{\|\{v_x^-\|, \|v_y^-\|\}, \\
\|v_z^+\| = C_v \|v_z^-\|, \\
\end{cases}
\label{eq:fit_eqs}
\end{equation}
where $\mathbf{v}$ and $\mathbf{a}$ are estimated from the first and second derivatives of $p(t)$, respectively. Using this model, we apply explicit step-by-step time integration to predict the ball’s future position and velocity, initialized with the estimates from Sec.~\ref{subsec:state_estimation}. Given a predefined virtual hit plane~\cite{mulling2010biomimetic, mulling2011biomimetic, kocc2018online} at $x = -1.37\,\mathrm{m}$, the corresponding hitting time and position can then be computed.

\begin{figure}
    \centering
    \includegraphics[width=1\linewidth]{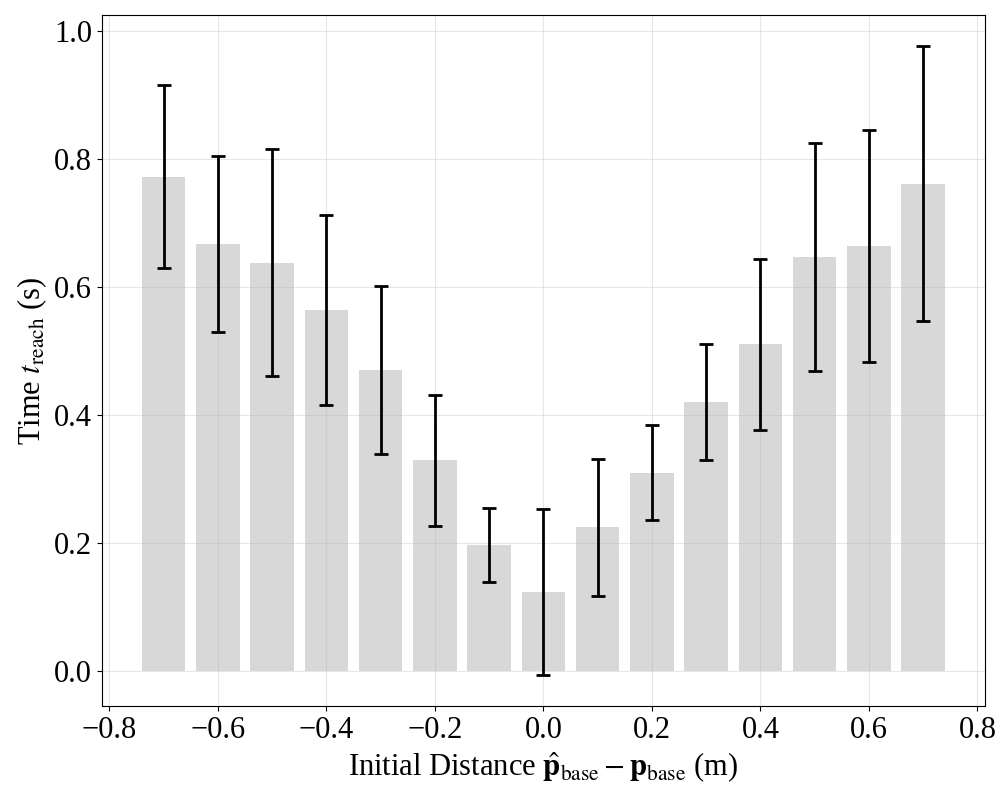}
    \caption{\textbf{Agility evaluation of the WBC policy.} Based on 943 successful simulated trials (94.3\% success rate), when the initial distance is within 0.75\,m, the target can be reached in under 0.8\,s on average, which is faster than the strike time of 0.86\,s. Error bars denote standard deviation across trials.}
    \label{fig:agility_test}
\end{figure}

\subsection{Racket-ball Interaction}
In addition to making contact with the ball, our objective is to successfully return it, which requires determining the racket’s orientation and velocity at the moment of impact. We assume that, at impact, the racket plane is perpendicular to its velocity vector. Unlike approaches that aim to precisely control the landing position, our goal is to ensure a valid return. Thus, we adopt a simplified post-impact flight model and racket-ball interaction model.

During racket-ball contact, we assume a coefficient of restitution $C_r$ along the normal to the racket surface and neglect tangential friction. After impact, the ball is assumed to be subject only to gravity during flight. Given the desired landing position on the table $\hat{\mathbf{p}}_l$, the desired hitting position $\hat{\mathbf{p}}_{\mathrm{racket}}$, and a predefined flight time from hitting to landing $\Delta t$, the desired outgoing ball velocity $\mathbf{v}_o$ is computed as:
\begin{equation}
\mathbf{v}_o=\dfrac{\hat{\mathbf{p}}_l-\hat{\mathbf{p}}_\mathrm{racket}}{\Delta t}+\frac{1}{2}\mathbf{g}\Delta t,
\end{equation}
where $\hat{\mathbf{p}}_l$ is set to the center of the opponent’s side of the table. Based on the desired outgoing velocity $\mathbf{v}_o$ and the predicted incoming velocity $\mathbf{v}_i$, the desired racket velocity $\hat{\mathbf{v}}_{\mathrm{racket}}$ can be computed as:
\begin{equation}
\hat{\mathbf{v}}_{\mathrm{racket}} = \dfrac{\mathbf{v}_o\cdot\mathbf{u}+C_r \mathbf{v}_i\cdot\mathbf{u}}{1+C_r}\mathbf{u},
\end{equation}
where $\mathbf{u}=\dfrac{\mathbf{v}_o - \mathbf{v}_i}{\|\mathbf{v}_o - \mathbf{v}_i\|}$ is the unit vector in the direction of $\mathbf{v}_o - \mathbf{v}_i$. 
The desired racket velocity, together with the predicted hitting position and time, is then passed to the learning-based whole-body controller, which generates the corresponding robot motion.

\section{Learning-based Whole-Body Controller}
The model-based planner predicts the desired racket striking position $\hat{\mathbf{p}}_{\mathrm{racket}}$, velocity $\hat{\mathbf{v}}_{\mathrm{racket}}$, and timing $t_{\mathrm{strike}}$. 
These predictions are provided to the learning-based Whole Body Controller (WBC), which generates the corresponding whole-body motions for the humanoid. We train the WBC policy $\pi_\text{WBC}$ in Isaac Lab~\cite{mittal2023orbit} and deploy it to the real robot in a zero-shot manner. The policy is trained end-to-end using the model-free reinforcement learning algorithm PPO~\cite{schulman2017proximal}. The joint PD gains are set heuristically following~\cite{liao2025beyond}.

\begin{table}
\centering
\caption{Observation spaces for the policy (actor) and critic. The critic receives additional privileged information during training.}
\begin{tabular}{lcc}
\toprule
\textbf{Observation} & \textbf{Actor} & \textbf{Critic} \\
\midrule
Base angular velocity ($\bm{\omega}_{\mathrm{base}}\in\mathbb{R}^3$) & \checkmark & \checkmark \\
Projected gravity vector ($\mathbf{g}_{\mathrm{base}}\in\mathbb{R}^3$) & \checkmark & \checkmark \\
Base forward vector ($\mathbf{e}_{\mathrm{base},x}\in\mathbb{R}^2$) & \checkmark & \checkmark \\
Target base position ($\hat{\mathbf{p}}_{\mathrm{base},xy}-\mathbf{p}_{\mathrm{base},xy}\in\mathbb{R}^2$) & \checkmark & \checkmark \\
Target racket position ($\hat{\mathbf{p}}_{\mathrm{racket}}\in\mathbb{R}^3$) & \checkmark & \checkmark \\
Target racket velocity ($\hat{\mathbf{v}}_{\mathrm{racket}}\in\mathbb{R}^3$) & \checkmark & \checkmark \\
Time to strike ($t_{\mathrm{strike}}\in\mathbb{R}$) & \checkmark & \checkmark \\
Joint positions ($\mathbf{q}\in\mathbb{R}^{29}$) & \checkmark & \checkmark \\
Joint velocities ($\dot{\mathbf{q}}\in\mathbb{R}^{29}$)& \checkmark & \checkmark \\
Previous action ($\mathbf{a}_{\mathrm{last}}\in\mathbb{R}^{29}$)& \checkmark & \checkmark \\
Base linear velocity ($\mathbf{v}_{\mathrm{base}}\in\mathbb{R}^3$) & -- & \checkmark \\
Bodies' pose ($\mathbf{T}_\mathcal{B}\in\mathbb{R}^{7|\mathcal{B}|}$) & -- & \checkmark \\
Time left in current episode ($t_{\mathrm{left}}\in\mathbb{R}$) & -- & \checkmark \\
\makecell[l]{Reference joint positions and velocities \\ ($[\hat{\mathbf{q}}, \hat{\dot{\mathbf{q}}}]\in\mathbb{R}^{58}$)} & -- & \checkmark \\
\bottomrule
\end{tabular}
\label{tab:obs_spaces}
\end{table}

\subsection{Human Motion References}
\label{subsec:hmr}

We extend the idea of generating \textit{instantaneous} racket motion close to a ``comfort posture" in \cite{mulling2010biomimetic, mulling2011biomimetic} to generating \textit{continuous} humanoid motion close to two swinging references: \textbf{forehand} and \textbf{backhand}. Note that while \cite{mulling2013generalizestrike} synthesizes a mixture of motor primitives controller from kinesthetic teach-ins, we use video-based demonstrations. First, we record a video clip of a human performing the swing. A corresponding SMPL \cite{loper2023smpl} motion clip is reconstructed from the video using GVHMR~\cite{shen2024world}, and then re-targeted to the humanoid robot using GMR~\cite{ze2025gmr, ze2025twist}. The resulting motion clip contains base pose and joint positions at 30\,Hz.

Following the approach of BeyondMimic~\cite{liao2025beyond}, we enhance the motion for better tracking. We first interpolate the base pose and joint positions from 30\,Hz to 50\,Hz, matching the control frequency of the policy. Then, base linear velocities, angular velocities, and joint velocities are computed using central differencing. Using forward kinematics, we obtain the pose $\mathbf{T}_{b}$ and twist $\mathbf{V}_{b}$ for each body $b \in \mathcal{B}$. Since we only track the motion of the upper body, $\mathcal{B}$ contains only the bodies above the pelvis (pelvis, torso, and arms). The pelvis is chosen as the anchor body $b_{\mathrm{anchor}}$, and the desired pose of other bodies is computed as in~\cite{liao2025beyond}. After processing, each motion clip contains 94 frames (1.88\,s) with the striking occurring at the $43^{\mathrm{nd}}$ frame (0.86\,s).

\begin{figure}
    \centering
    \includegraphics[width=1\linewidth]{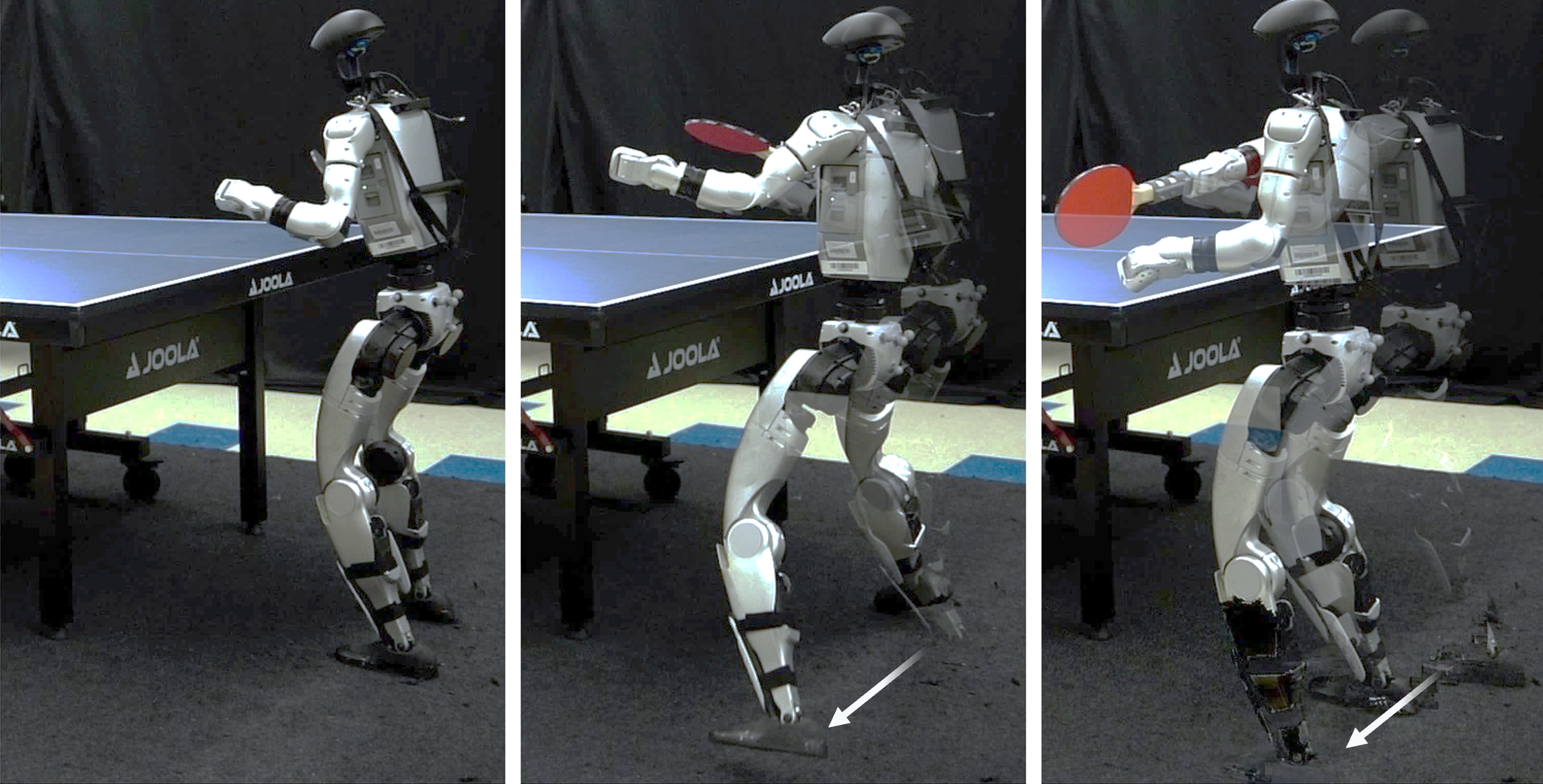}
    \caption{\textbf{Real-world rapid reaching motion.} The whole-body control policy enables agile reaching motions, allowing the robot to swiftly transition from the right side of the table to the left while maintaining balance and successfully striking the ball.}
    \label{fig:side_jump}
\end{figure}

\subsection{Markov Decision Process (MDP) Setting}

\subsubsection{Separate Commands for Base and Racket}
Instead of using the global racket position and velocity at the striking time as commands~\cite{ma2025learning}, we separate the commands for the base and the racket to improve training efficiency. The first command $\hat{\mathbf{p}}_{\mathrm{base},xy}$ specifies the desired base position in the world frame, encouraging the robot to arrive at the target location in time. The second command $[\hat{\mathbf{p}}_{\mathrm{racket}}, \hat{\mathbf{v}}_{\mathrm{racket}}]$ specifies the racket position relative to the base and the racket velocity (both expressed in the world frame). The desired base orientation is always set to face forward.

To enable the policy to perform consecutive strikes and switch between swing types, each episode lasts 10\,s. After completing a swing, we uniformly sample the next swing type (forehand or backhand) and randomly sample the racket target position, racket target velocity, and base target position at the striking time, conditioned on the swing type. The striking plane is fixed at 0.4\,m in front of the robot, so only the $y$ and $z$ coordinates of the racket target position are sampled. The target regions for forehand and backhand are defined to be non-overlapping.

\subsubsection{Reward Functions}
To generate motions that both resemble the reference motions and track the given commands, we follow~\cite{peng2018deepmimic} and define the total reward as:
\begin{equation}
    r = w_i r_i + w_g r_g + w_r r_r,
\end{equation}
where $r_i$ encourages imitation of the upper body reference motion, $r_g$ rewards tracking of the commanded goals, and $r_r$ provides regularization. $w_i$, $w_g$, and $w_r$ are the corresponding reward weights.

Both $r_i$ and $r_r$ are dense rewards that are applied throughout the entire episode.  
In contrast, $r_g$ contains several sparse terms with relatively high weights.  
For instance, the tracking rewards for the racket position, velocity, and orientation are only activated during a short window around the hitting time, while the base position tracking reward is activated only before the strike, enabling the policy to prepare for transitioning to the next target position after hitting.

\begin{figure}
    \centering
    \includegraphics[width=1\linewidth]{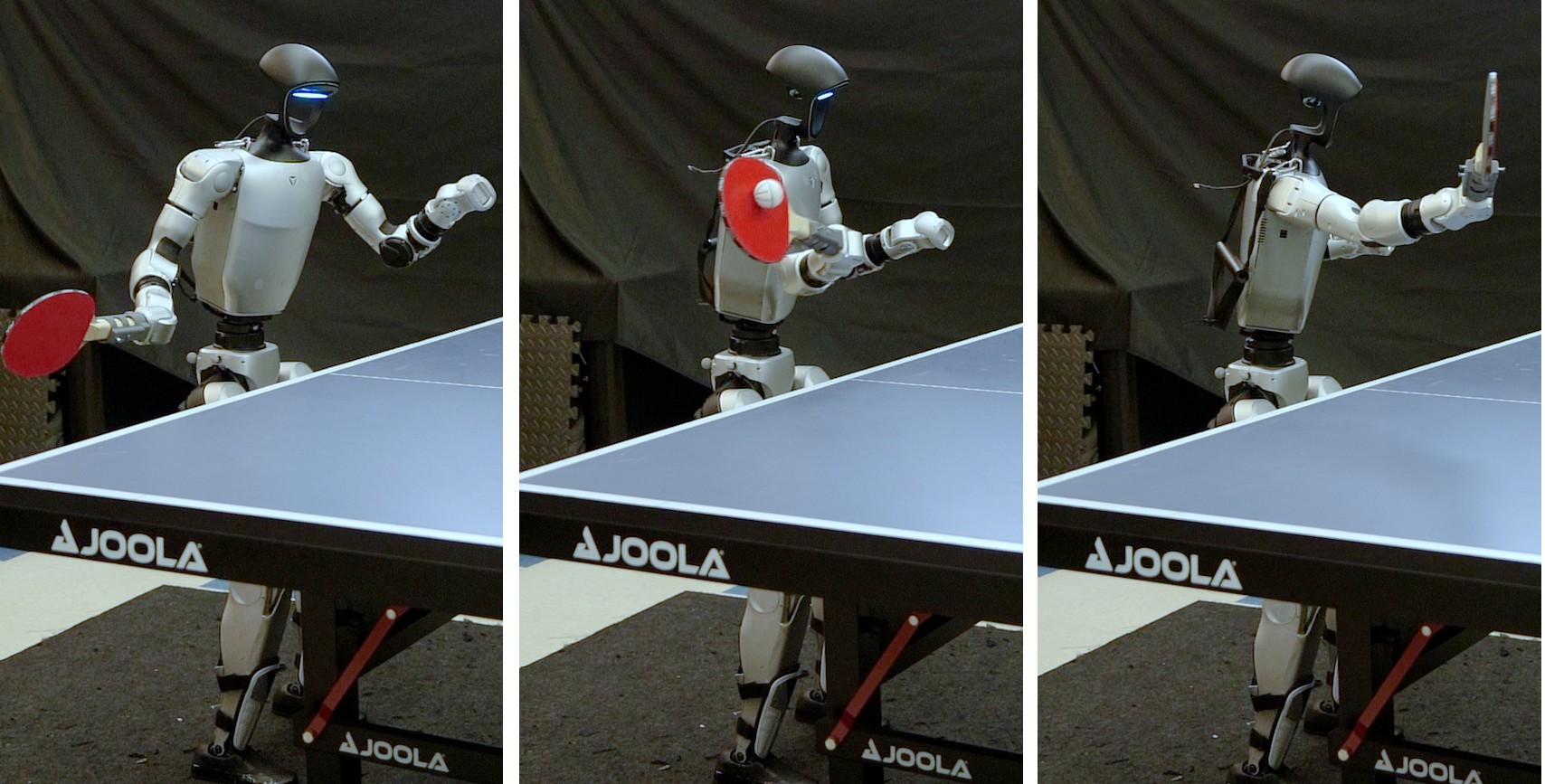}
    \caption{\textbf{Real-world human-like striking motion.} The whole-body control policy generates human-like striking motions, including coordinated waist rotation during hits, which mimics the way humans play table tennis.}
    \label{fig:hit_motion}
\end{figure}

\subsubsection{Asymmetric Actor-Critic}

To provide the critic with additional information unavailable to the policy at deployment, we adopt an asymmetric actor-critic framework for training~\cite{pinto2018asymmetric} (Table~\ref{tab:obs_spaces}).  
For example, to improve reference tracking, we augment the critic’s observations with the robot body poses $T_{\mathcal{B}}$, which facilitate more accurate return estimation. 
Since several terms of $r_g$ are sparse, the episodic return depends on the number of strikes remaining within an episode.  
Therefore, we additionally provide the critic with the time left in the current episode, $t_{\mathrm{left}}$. Both the actor and critic are implemented as multi-layer perceptrons (MLPs) with three hidden layers of sizes 512, 256, and 128.

During deployment, the motion capture system provides the robot base’s position and orientation, which are used to construct $\mathbf{e}_{\mathrm{base},x}$ and $\mathbf{p}_{\mathrm{base},xy}$ in the observation vector. Based on $\mathbf{p}_{\mathrm{base},xy}$ and the predicted racket position $\hat{\mathbf{p}}_{\mathrm{racket}}$, we heuristically determine whether a forehand or backhand strike should be used. This binary variable is then employed to compute the desired base position $\hat{\mathbf{p}}_{\mathrm{base},xy}$. Note that this variable is not included in the policy observations, and it serves only to assist in computing $\hat{\mathbf{p}}_{\mathrm{base},xy}$.

\section{Results}
In our experiments, we aimed to answer three key questions:  
(a) How accurate is the model-based prediction system?  
(b) How agile is the whole-body control policy designed for table tennis?  
(c) When integrated, how effectively can the system return the ball and even play against human or humanoid opponents in the real world?

\begin{figure*}
    \centering
    \includegraphics[width=\linewidth]{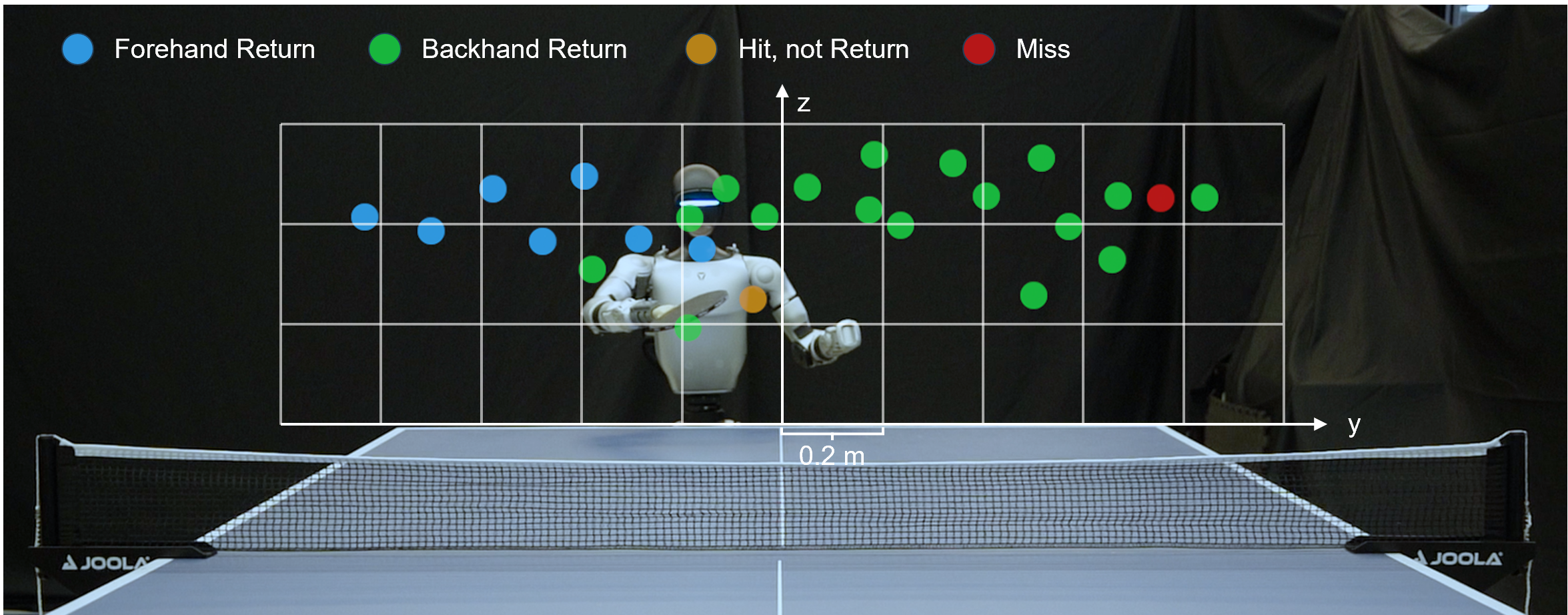}
    \caption{\textbf{Return performance evaluation.} The results are displayed on a virtual hit plane, with each block side representing 0.2\,m. Of the 26 incoming balls distributed across the plane, the robot successfully returned 24 (blue: forehand strokes, green: backhand strokes), while 1 hit did not result in a return (orange) and 1 was a complete miss (red). This corresponds to a 96.2\% hit rate and a 92.3\% return rate.}
    \label{fig:succ_rate}
\end{figure*}

\subsection{Model-based Planner}

To evaluate the performance of the model-based planner, we collect 20 ball trajectories and compute the prediction errors for both the hitting position and strike time (Fig.~\ref{fig:prediction_error}). The errors decrease as the strike approaches, reaching zero at the moment of contact. This trend aligns with the intuition that longer prediction horizons accumulate larger errors. At 0.5\,s before the strike, the position prediction error falls below the critical threshold of 7.5\,cm, corresponding to the racket’s radius. At 0.3\,s before the strike, the time prediction error drops below 20 ms, equivalent to one control step of the policy. At 0.1\,s before the strike, both position and time errors reach minimal level. The consistently low error across the entire prediction process provides a stable command interface for the WBC policy and establishes a foundation for the overall accuracy of the system.

\subsection{Whole-Body Control}

We examine the relation between the initial distance from the desired base position $\hat{\mathbf{p}}_{\mathrm{base}}$ to the actual base position $\mathbf{p}_{\mathrm{base}}$ at the moment a new command is sampled, and the time required to reach within 1\,cm of $\hat{\mathbf{p}}_{\mathrm{base}}$ to evaluate the agility of the whole-body control policy. We collect 1000 roll-outs in simulation and discard 57 cases where the 1\,cm threshold is not reached, resulting in 943 valid data points and a success rate of 94.3\%. As shown in Fig.~\ref{fig:agility_test}, when the initial distance is less than $0.75$\,m, nearly all trials converge to within 1\,cm of $\hat{\mathbf{p}}_{\mathrm{base}}$ in less than 0.8\,s, which is shorter than the typical duration from command issuance to strike (0.86\,s), as described in Sec.~\ref{subsec:hmr}. This indicates that, in almost all cases, the robot successfully reaches the desired base position before the striking motion is completed. Moreover, convergence time increases monotonically with the initial distance, as expected. The statistical distribution is highly symmetric, with comparable convergence times for displacements in both the left and right directions. The results demonstrate the agility of the WBC policy in simulation.

In the real world, we observe that the policy consistently exhibits agile motions. As shown in Fig.~\ref{fig:side_jump}, the robot initially stands on the right side of the table. When the ball is directed toward the left side, it reacts quickly and, with a single step, moves across to return the ball rapidly. This prompt, one-step motion is enabled by commanding the base position. In contrast, when using velocity commands as in prior WBC approaches~\cite{cheng2024expressive}, the robot consistently performs slower, multi-step lateral movements. For the upper-body arm swing, training with human motion references produces striking behaviors that closely resemble human motions, including waist rotation during the hit, as demonstrated in Fig.~\ref{fig:hit_motion}.

\subsection{Real-World Experiments}
When we deploy our system, we throw 26 balls toward the robot, with their projections on the virtual hit plane spanning a wide area. The robot achieves 24 successful returns, misses 1 return after a hit, and completely misses 1 ball, corresponding to a 96.2\% hit rate and a 92.3\% return rate (Fig.~\ref{fig:succ_rate}). We also observe that the robot tends to use forehand strokes for balls incoming at $y < 0$ and backhand strokes for balls incoming at $y > 0$, which is consistent with human table tennis play.

Moreover, our humanoid can engage in extended play against human opponents, achieving rallies of up to 106 consecutive shots. This rally ends when the humanoid hits the ball into the net. Such a rally length exceeds that of casual human play, indicating that the humanoid not only tracks and returns the ball reliably but also maintains balance and readiness across successive strokes. The humanoid can even return human smashes with only 0.42\,s of reaction time from the opponent’s hit to the robot’s return. Furthermore, when two humanoids equipped with the same policy face each other, they can sustain continuous rallies in a fully autonomous match setting (Fig.~\ref{fig:teaser}).

\section{Discussion}
\label{disc-sec}
\subsection{System Design}
Our system adopts two orthogonal combinations: (1) a high-level planner with a low-level controller, and (2) a model-based method with a learning-based method. The first combination modularizes the architecture, separating long-horizon prediction and planning from short-horizon whole-body control.
This modularization enables the two modules to be independently evaluated and progressively improved, for instance, by quantifying the planner’s prediction accuracy (Fig.\ref{fig:prediction_error}) and the controller’s agility (Fig.\ref{fig:agility_test}).

The second combination leverages the strengths of model-based and learning-based methods, while mitigating their respective shortcomings.
In table tennis setting, where rewards are sparse and delayed, end-to-end RL often struggles with exploration and suffers from low sample efficiency. 
Purely model-based approaches, in contrast, rely on highly accurate dynamics and perception models, which are difficult to obtain for humanoids with many degrees of freedom and frequent ground contacts. 
Our hierarchical design bridges this gap: it improves sample efficiency, increases robustness to perception errors, and adapts effectively to real-world conditions. Consequently, the system can react in real time to the sub-second dynamics of the game, and simultaneously generate agile strike motion and footwork, achieving a high rate of successful returns.

\subsection{Limitations}
Despite demonstrating promising results in humanoid table tennis, several limitations remain.

\textbf{Virtual hitting plane.} The current system assumes a fixed hitting plane at the end of the table, which constrains striking strategies and reduces effectiveness against very short or deep balls. As a result, when playing against the humanoid, a human opponent must avoid hitting overly short balls that the robot cannot reach. Relaxing this assumption could enable more diverse contact points and improve table coverage.

\textbf{External motion capture.} Ball position and robot base pose are provided by a motion capture system, restricting deployment to controlled environments. Incorporating vision-based sensing would alleviate this dependency and allow operation in more natural and diverse settings.

\textbf{Spin handling and stroke repertoire.} The system assumes negligible spin and relies on a flat push to return the ball. Professional-level play, however, involves heavy spin and diverse strokes, e.g., top-spin loops, back-spin chops, side-spin blocks. Extending the system to perceive spin and generate appropriate counter-strokes would bring humanoid performance closer to expert human play.

\subsection{Future Work}
\textbf{Multi-agent training and serving.} In our current humanoid-humanoid experiments, both robots were equipped with the same policy without any joint training of different policies. Even under this simple setting, they were able to sustain a table tennis game. Future work could explore explicit multi-agent training frameworks to further improve competitiveness. Another point is that our robots are not yet capable of serving; in both humanoid-human and humanoid-humanoid settings, a human player is required to initiate the rally. Enabling autonomous serving thus represents another important direction for future work.

\textbf{Learning to play against skilled opponents.} We believe that we have established a strong baseline for the making of table tennis shots by humanoid robots. In the future, we expect these humanoids to learn to play against skilled opponents. The additional characteristics that are required to play against skilled opponents include learning their characteristics by watching them play (strategic learning), and then adapting the stroke making in real time to specific characteristics of the opponent (tactical learning). We have begun a program to determine the intent of the opponent in \cite{etaat2025latte}. We hope to integrate these methods along with recent advances in Markov games \cite{KMS2025} to have championship-quality humanoid robot players. 

\section{Conclusions}
We have presented a hierarchical system that enables humanoid robots to play table tennis in the real world. By combining a model-based planner for accurate ball trajectory prediction with a reinforcement learning-based whole-body controller, our approach achieves agile, human-like striking motions under sub-second reaction times. 
Real-world experiments demonstrate high return success rates, natural forehand/backhand strategies, and rallies of up to 106 consecutive shots against human opponents. Furthermore, we show that two humanoid robots can autonomously sustain rallies against each other, highlighting the robustness and generality of our framework. 
These results advance humanoid control toward more agile, interactive, and human-level behaviors. Eventually we hope to have humanoids play championship calibre table tennis against skilled opponents. 







\section*{ACKNOWLEDGMENT}

We would like to thank Zhaoming Xie and Takara Truong for their insightful discussions, and Toby Yegian for his assistance with the motion capture system. This work was performed in the FHL Vive Center for Enhanced Reality and was supported in part by The Robotics and AI Institute, NSF CMMI-2140650, and by the program ``Design of Robustly Implementable Autonomous and Intelligent Machines (TIAMAT)", Defense Advanced Research Projects Agency award number HR00112490425.


{
\balance
\bibliographystyle{IEEEtran}
\bibliography{bibtex}
}

\end{document}